\documentclass[a4paper]{article}

\usepackage{sods_2023}[final]
\usepackage{times}            
\usepackage[english]{babel}   
\usepackage[ansinew]{inputenc}
\usepackage[T1]{fontenc}      
\usepackage{amsmath,amssymb}
\usepackage{graphicx}
\usepackage[colorlinks=false,pdfborder={0 0 0}]{hyperref}
\usepackage{units}

\usepackage{graphicx}
\usepackage{booktabs}
\usepackage{parskip}
\usepackage[toc,page]{appendix}
\nolinenumbers

\begin{document}

\title{Efficient data selection employing Semantic Similarity-based Graph Structures for model training}



\maketitle

\begin{abstract}
Recent developments in natural language processing (NLP) have highlighted the need for substantial amounts of data for models to capture textual information accurately. This raises concerns regarding the computational resources and time required for training such models. This paper introduces \textbf{SE}mantics for data \textbf{SA}liency in \textbf{M}odel performance \textbf{E}stimation (\textbf{SeSaME}). It is an efficient data sampling mechanism solely based on textual information without passing the data through a compute-heavy model or other intensive pre-processing transformations. The application of this approach is demonstrated in the use case of low-resource automated speech recognition (ASR) models, which excessively rely on text-to-speech (TTS) calls when using augmented data. SeSaME learns to categorize new incoming data points into speech recognition difficulty buckets by employing semantic similarity-based graph structures and discrete ASR information from homophilous neighbourhoods through message passing. The results indicate reliable projections of ASR performance, with a $93\%$ accuracy increase when using the proposed method compared to random predictions, bringing non-trivial information on the impact of textual representations in speech models. Furthermore, a series of experiments show both the benefits and challenges of using the ASR information on incoming data to fine-tune the model. We report a $7\%$ drop in validation loss compared to random sampling, $7\%$ WER drop with non-local aggregation when evaluating against a highly difficult dataset, and $1.8\%$ WER drop with local aggregation and high semantic similarity between datasets.
\end{abstract}

\section{Introduction}


Cutting-edge advancements have emerged across several areas of artificial intelligence, including large language models \cite{Yang2019XLNet}, multi-modal and context-aware models \cite{Li2019VisualBert}, conversational AI \cite{brown2020GPT3} and vision transformers \cite{sharir2021VIT}. While they gain robustness and generalizability across tasks, they often become more computationally expensive and data-demanding. This leads to problems concerning the availability of trustworthy data \citep{chen2017labelleddata}, budget allocation, and environmental impact.

A ubiquitous example of a system requiring substantial quantities of data is automated speech recognition (ASR), for which large-scale training can significantly improve model performance \cite{long2019large-scale}.  Examples of commonly used benchmark datasets are VoxLingua107, containing speech segments extracted from YouTube videos that amount to $6628$ hours of data, and LibriSpeech, incorporating $960$ hours of audiobook data \citep{li2020librispeech}.
Training an ASR system has several challenges: (1) the computational workload and time required for processing audio data are costly, and (2) several low-resource languages lack annotated data. Data augmentation is one of the most common techniques to compensate for a low resource setting \cite{Wei2019DataAugmentationText, Park2019DataAugmentationSpeech}. However, it is not necessarily the case that adding vast quantities of synthetic data will proportionally improve the model's performance, while it does add to the computational workload \cite{sun2017data_in_dl}.

We present \textbf{SE}mantics for data \textbf{SA}liency in \textbf{M}odel performance \textbf{E}stimation (\textbf{SeSaME}), a novel graph-based approach to finding salient training instances using semantic similarity between data points. Specifically, we focus on the ASR task and investigate if, given a set of textual utterances, we can select a subset for fine-tuning an ASR system and achieve better performance as if fine-tuning on a random sample from the same dataset. \emph{Our intuition is to use the measured model to infer its evaluation performance on a new dataset through label recovery on the utterance level based on semantic similarity between the new sentences and the observed data.} Efficient data sampling brings two advantages: (1) compute benefits by reducing speech synthesis calls and (2) lower carbon footprint as a result of more efficient training. 
We propose an approach to estimating ASR performance using semantic similarity-based graph neural networks and leverage the salient data points for fine-tuning an ASR system. In this paper, we answer the following research questions: (1) can we use semantic priors and graph-based structures to predict the performance of an ASR model? and (2) if so, how can we use the leveraged information to sample data points and fine-tune the ASR model?

Our key contributions are the following:
\begin{itemize}
    \item We propose SeSaME, a novel approach to modelling the performance of a model in discrete space using graph-based semantic similarity using textual data alone.
    \item We leverage known model performance to efficiently sample new data for fine-tuning.
    \item We show that by incorporating an attention mechanism, our proposed sampling procedure achieves a $7\%$ WER improvement compared to the baseline.
\end{itemize}

The remainder of the paper is organized as follows: In Section \ref{section:methodology} we formalize the approach of using textual semantic similarity graph structures to predict ASR performance. Section \ref{section:experimental_setup} presents the experimental setup for training and fine-tuning the ASR and GNN models, repectively. Section \ref{section:results} discusses the results and answers the research questions, while Section \ref{section:conclusion} lays down the conclusions of the experiments and proposes future research directions.

\section{Related Work}

In this section, we outline the definitions needed for formulating our approach.

\textbf{Graph Neural Networks} Graph neural network (GNN) architectures emerged as powerful tools for processing and exchanging node, edge and structure information \citep{cui2020GNN} through message passing. Message passing (MP) updates each node representation based on its 1-hop neighbourhood information. MP layers differ by choice of the aggregation function over the 1-hop neighbourhood. Depending on the architecture, the aggregation can take different forms. We will study and compare the impact of local aggregation, i.e., GCN \citep{kipf2016semisupervised}, GIN \citep{xu2019powerful}, and GraphSAGE \citep{hamilton2017inductive}, with non-local aggregation, i.e., GAT \citep{velickovic2017graphattention}.

\textbf{Label Recovery} Assume a graph $\mathcal{G} = (\mathcal{V}, \mathcal{E})$, where $\mathcal{V}$ is the set of vertices and $\mathcal{E}$ is the set of edges representing connections between nodes. The \emph{label recovery task} is the problem of inferring missing labels for a set of nodes $\mathcal{V'} \subseteq \mathcal{V}$ from available information, i.e., known labels of nodes $\mathcal{V} \setminus \mathcal{V'}$. The labels can be either discrete, i.e., in which case the task is a classification problem, or continuous, i.e., in which case the task is a regression problem.

\textbf{Homophily} In the context of social networks, homophily has been expressed as the tendency of nodes to connect with others that exhibit similar characteristics. In contrast, heterophily is the tendency of nodes to gather into scattered groups. It has been shown that the degree of homophily directly impacts the performance of graph models in a label recovery task \cite{ke2019labelrecovery}, i.e. higher homophily leads to better performance. We use the Schelling model \citep{pollicott2001schelling} to assess the homophily of sampled neighbourhoods in the graph structure for different ASR utility functions. Accordingly, we choose the utility that exhibits a higher degree of homophily for optimizing the performance of the graph model on the label recovery task.

\section{Methodology}\label{section:methodology}

\begin{figure*}[!htb]
 \centering
 \includegraphics[width=\columnwidth]{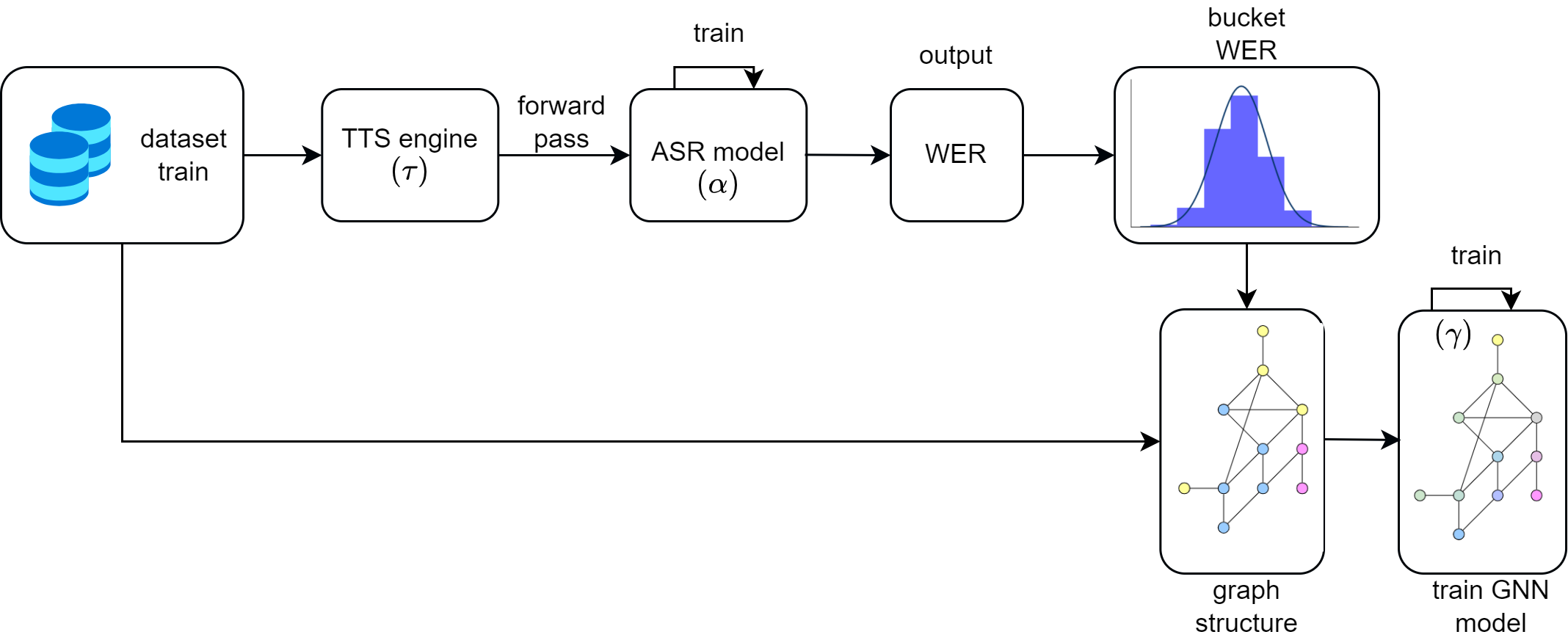}
 \caption{Train pass from left to right: Given dataset $\mathcal{D}_\text{train}$; Pass it to the TTS engine $\tau$ for transforming textual sentences into synthetic audio; Train ASR model $\alpha$; Compute WER on the predictions of $\alpha$; Bucket the observed WER into seven ordinal classes; Create a graph structure $\mathcal{G} = (\mathcal{V}, \mathcal{E})$ with BERT embeddings as node features, bucketed WER as labels, and edges between semantically similar nodes; Train GNN $\gamma$ on the label recovery task.}
 \label{fig:entire_framework_1}
\end{figure*}

\begin{figure*}[!htb]
 \centering
 \includegraphics[width=1\columnwidth]{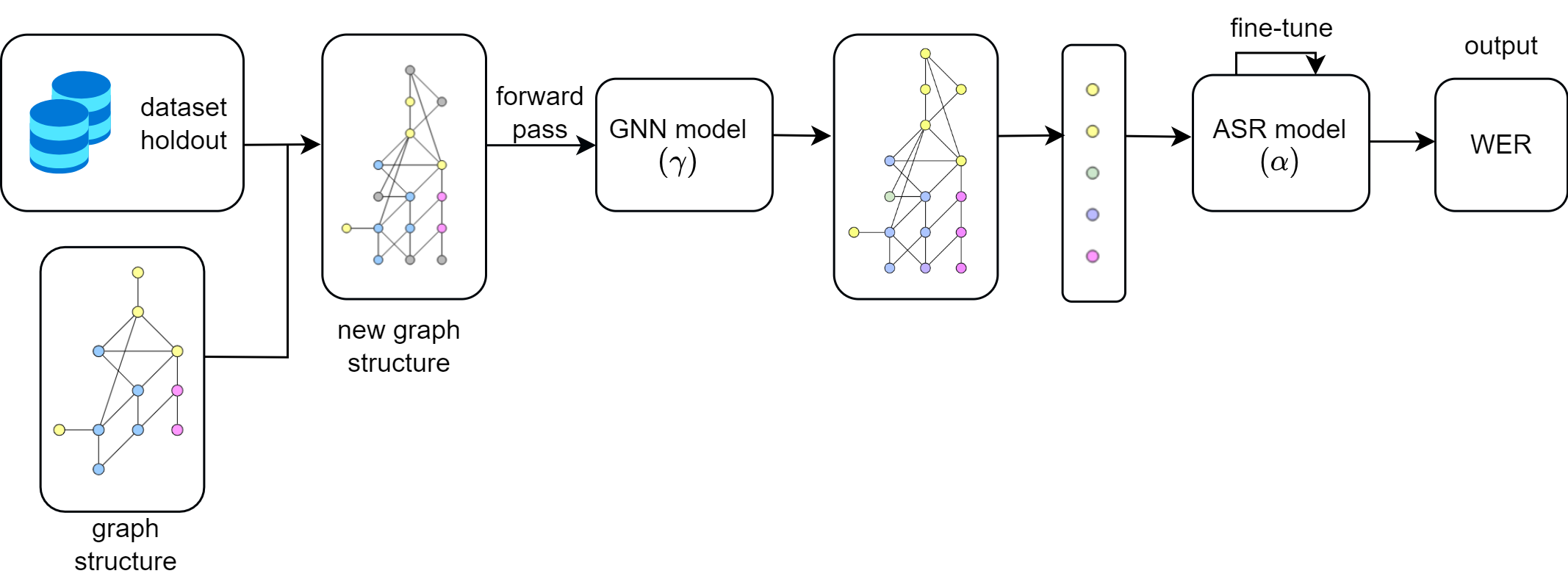}
 \caption{Fine tune pass from left to right: Given new dataset $\mathcal{D}_\text{holdout}$; Add it to $\mathcal{G}$ to create a new graph structure $\bar{\mathcal{G}} = (\bar{\mathcal{V}}, \bar{\mathcal{E}})$; Make forward pass to the pre-trained GNN to retrieve labels for $\mathcal{D}_\text{holdout}$; Sample the nodes predicted as highly difficult points; Fine-tune $\alpha$ with the sampled data; Get WER performance and compare with baselines.}
 \label{fig:entire_framework_2}
\end{figure*}

In this section, we explain SeSaME, our semantic graph-based sampling approach, considering the following ASR use case: assume access to a high-cost pre-trained ASR model $\alpha$; we want to estimate its performance on a new dataset without explicitly using the data, i.e., without making a forward pass; the only available information is the prior training data points and their ASR performance. We formalize and split our approach in two parts: (1) the train pass (see Figure \ref{fig:entire_framework_1}) constructs and trains a semantic similarity graph using the available training data and its ASR evaluation metrics, and (2) the fine tune pass (see Figure \ref{fig:entire_framework_2}) uses the graph structure for mapping incoming data points to ASR performance, and uses the leveraged information for sampling a subset of the incoming data for further fine-tuning the ASR model. A summary of the notation used throughout this section is presented in Appendix \ref{section:appendix}.

\subsection{Train Pass}

\textbf{ASR Training } Consider a textual dataset $\mathcal{D}_\text{train}$ and a text-to-speech (TTS) engine $\tau$ that receives as input $\mathcal{D}_\text{train}$ and generates its corresponding synthetic audio. The audio synthesis is then used for training an ASR model $\alpha$. The ASR model predicts a hypothesis for each data point and is evaluated against its reference (ground truth) sentence using Word Error Rate (WER), which measures the percentage of incorrectly recognized or substituted words in the sentence.

Our intuition is that we can infer the model's performance on a new dataset $\mathcal{D}_\text{holdout}$ using the observed WER and the semantic similarity of $\mathcal{D}_\text{holdout}$ with $\mathcal{D}_\text{train}$. The approach is inspired by the RHO-LOSS \citep{mindermann2022RHOLOSS} utility function presented in Equation \ref{eq:rho}, which has been proven efficient in sampling task-relevant, non-redundant and non-noisy data for training on a wide range of datasets and tasks. We will use the first term of the function to define a label for each sentence in $\mathcal{D}_\text{train}$:

\begin{equation}\label{eq:rho}
    \mathcal{L}[y|x] = \operatorname{argmax}_{(x, y) \in B_t} \mathcal{L}[y|x;\mathcal{D}_\text{train}] - \mathcal{L}[y|x;\mathcal{D}_\text{holdout}].
\end{equation}

where $x$ represents the input waveform fed to the ASR model, $y$ is the prediction (hypothesis), $\mathcal{L}[y|x;\mathcal{D}_\text{train}]$ is the training loss of the model, and $\mathcal{L}[y|x;\mathcal{D}_\text{holdout}]$ is the loss of the model when trained on a smaller adjacent dataset, called the holdout dataset;

We aim to choose a labelling utility function $L[y|x;\mathcal{D}]$ that has the following properties: it is suitable for evaluating the ASR model, it is representative of homophilous relationships between input sentences, and it can be discretized into ordinal labels. We have experimented with both WER and CTC loss as utility functions; we chose to use WER as it exhibits a higher degree of homophily in the graph. 

The WER metric is defined as follows:

\begin{equation}\label{eq:wer}
    WER = \frac{S+D+I}{N},
\end{equation}

Equation \ref{eq:wer} can be interpreted as the percentage of incorrectly substituted, deleted and inserted words, denoted as $\{S, D, I\}$, compared to the total number of words $N$ in the reference, where $N = S + D + C$. The typical value for WER is $w_i \in [0, 1]$, however, it can exceed the upper bound when the prediction (hypothesis) is longer than the ground truth sentence (reference). A lower WER indicates better performance.

However, WER has one nontrivial disadvantage when using it in an ordinal regression task: it is a continuous variable. To mitigate this issue, we discretise WER into $k$ buckets according to their distribution in $\mathcal{D}_\text{train}$, bringing WER from a continuous to a discrete space. Each sentence in $\mathcal{D}_\text{train}$ is mapped to a WER value which is associated with one of the defined classes to create a label.

\textbf{Graph Creation } We can construct an undirected weighted graph $\mathcal{G} = (\mathcal{V}, \mathcal{E})$ with nodes $\mathcal{V}$ and edges $\mathcal{E}$ using $\mathcal{D}_\text{train}$ and the inferred WER labels as follows:

\begin{itemize}
\item Each node $v \in \mathcal{V}$ is associated with the textual representation of one single data point from $\mathcal{D}_\text{train}$. The textual representations are modelled as BERT embeddings \citep{devlin2018bert}. We use the BERT base uncased model.
\item Two utterances $\{u, v\} \subseteq \mathcal{V}$ are connected through an edge $(u, v) \in \mathcal{E}$ iff the semantic similarity between them exceeds a configurable threshold. The similarity between any pair of sentences is calculated on the node embeddings and not the waveform.
\end{itemize}
Computing semantic similarity in a large, fully connected graph is computationally infeasible. As an alternative, we apply approximate nearest neighbours (ANN) search to connect edges in the graph with cosine similarity as edge weights \cite{hajeby2011NAA}.

Moreover, since the objective of this study is to predict ASR performance without making any TTS calls for converting augmented textual data into audio samples, waveform features are not employed in the graph creation process.

\textbf{Graph Neural Networks } We aim to learn a mapping between textual utterances and their WER labels, for which we train a GNN model that takes as input $\mathcal{G} = (\mathcal{V}, \mathcal{E})$ and outputs a WER prediction for each node. There are two reasons for modelling this problem using graph structures: (1) aggregating information through message passing (our intuition is that similar utterances have a similar impact on the ASR model), and (2) using edge weights (the degree of similarity should have an impact in the message passing process).
The WER labels indicate how well the ASR model can map its equivalent audio waveform to the reference sentence. We formalize the problem as an ordinal regression task by defining the label encoding and loss functions:

\begin{itemize}
    \item Label Encoding: If a data point $x$ is mapped to a label $y$, it is automatically classified into all lower labels $(0, 1, 2, ... k-1)$. The target of input $x$ is $t = (1, 1, .., 1, 0, 0, 0)$ where $t_i$ with $0 \leq i \leq k-1$ is set to $1$ and all other elements to $0$.
    \item Loss Function: We use binary cross entropy as a loss function to differentiate distance magnitudes between predictions and real targets. It has two advantages: (1) treating each class as a binary classification task and (2) applying a logarithmic scale between classes, ensuring a higher loss for wrong predictions that are further apart from the ground truth label. 
\end{itemize}

\subsection{Fine Tune Pass}

Consider a new augmented dataset $\mathcal{D}_\text{holdout}$ which contains textual utterances for further fine-tuning the ASR model. Instead of passing it through the TTS engine $\tau$ and ASR model $\alpha$ as previously done with $\mathcal{D}_\text{train}$, the GNN mapping can be used for predicting the importance of individual entries in $\mathcal{D}_\text{holdout}$ on fine-tuning $\alpha$ by creating a new graph $\bar{\mathcal{G}} = (\bar{\mathcal{V}}, \bar{\mathcal{E}})$.

The textual representations of $\mathcal{D}_\text{holdout}$ are added to the existing graph $\mathcal{G} = (\mathcal{V}, \mathcal{E})$ by creating a node $u$ for each sentence in $\mathcal{D}_\text{holdout}$, computing its approximate nearest neighbors $\mathcal{V}_u \subset \mathcal{V}$, and adding edges from $u$ to its neighbors $v \in \mathcal{V}_u$. The edge weights are the cosine similarities between nodes. The labels of the holdout dataset are predicted by passing $\bar{\mathcal{G}}$ through the trained GNN and solving the label recovery task.

The inferred labels of $\mathcal{D}_\text{holdout}$ indicate how well the ASR model performs on new incoming data without including them in training or even processing their audio synthesis. 
We can use these projected WER labels to sample a subset of points from $\mathcal{D}_\text{holdout}$ for further fine-tuning $\alpha$.

\section{Experimental Setup}\label{section:experimental_setup}

In this section, we specify the data and models used, i.e., ASR and GNN architectures, training hyperparameters, and preliminary results that influenced modelling decisions.

\textbf{Data } We use the Common Voice Corpus 11.0 English dataset \citep{rosana2020CommonVoice} which is available for public use. It contains over $3$k hours of recorded audio sourcing from over $84$k different voices with varying accents, which include demographic metadata such as names, age and locations. Each data point contains an audio file sampled at $16$ kHz and its corresponding textual transcription. Several pre-processing steps have been taken, i.e. lowercasing and removing punctuation. Moreover, we eliminated time-stretching outliers with over 165K parameters and data points for which the audio file was unavailable. After pre-processing, the train, validation and test datasets have a size of $885$k, $15$k and $15$k data points. As described in Section \ref{section:methodology}, we create buckets out of the WER evaluation metric to create ordinal categorical labels for our dataset. The buckets are chosen so that the classes do not differ too much in size. We divide WER into seven classes by grouping all utterances with a WER $x_i \leq X_c$ where $X_c = [0.05, 0.1, 0.15, 0.2, 0.3, 0.5, 1]$. In the case when WER $x_i > 1$ we include the utterance into the last bucket.

\textbf{ASR model } We employ the wav2vec2 XLS-R-300M model for cross-lingual speech representation learning. It comprises a CNN feature extractor, a feature encoder generating multilingual speech units, and a transformer. It has been pre-trained using $435$k hours of unlabeled speech from open-source datasets such as VoxPopuli \citep{2022voxpopuli}, Common Voice \citep{2020commonvoice}, VoxLingua107 \citep{2021voxlingua} and BABEL \citep{2021babel}.
To use the XLS-R-300M architecture in an ASR setting, we drop the transformer and fine-tune it to solve a connectionist temporal classification (CTC) task. 
We train the model with Common Voice English for 20 epochs using a batch size of 8 utterances, Adam optimizer, learning rate of $3 \times 10^{-4}$, CTC loss function, and evaluate it with WER.

\textbf{GNN Models } During the experimental phase we discovered that the BERT embeddings are not sufficiently discriminative between classes, making it difficult for the GNN to learn a mapping from textual utterances to their WER labels. To mitigate this issue, we employ a feature embedding MLP trained with self-supervised contrastive loss \cite{prannay2020supervised_contrastive_loss} to push BERT embeddings closer together if assigned to the same class or further apart if assigned to different classes. The MLP has two linear layers that keep the BERT dimensionality consistent and one $\mathsf{relu}$ activation function. Using the MLP BERT embeddings, we deploy four GNN models:

\begin{itemize}
    \item \textbf{Graph Convolutional Network (GCN)}: GCN \cite{welling2017CGN} is a Graph Neural Network (GNN) architecture that applies a multilayer local aggregation function (MLA). It is an efficient and unbiased method of solving the label recovery problem in a graph structure.
    \item \textbf{Graph Isomorphism Network (GIN)}: GIN \cite{xu2019powerful} is a more flexible and expressive architecture than GCN, which also uses a local aggregation technique.
    \item \textbf{GraphSAGE}: GraphSAGE \cite{hamilton2017inductive} is an inductive learning GNN that learns an aggregation function to recognize both local information and global position. The graph is thus able to learn complex properties of the neighbourhoods, which leads to better generalization on unseen nodes.
    \item \textbf{GAT}: GAT \cite{velickovic201GAT} is a non-local aggregation GNN that makes use of masked self-attention layers to attribute attention weights to different neighbourhoods of a node in $\mathcal{G}$. GAT brings the advantage of not needing to rely on homophilous graphs.
\end{itemize}
Each GNN model consists of $4$ message passing layers with a hidden dimensionality reduction from $768$ to $128$ nodes and tanh as nonlinearity, one linear layer and a final sigmoid function for each class output node. The GNNs are trained for $2100$ epochs using drop edge \citep{yu2020dropedge} with probability $p=0.25$ to avoid over smoothing, Adam optimizer, weight decay $10^{-4}$, learning rate $10^{-3}$, binary cross entropy, and evaluated across accuracy and one frame agreement (OFA). The OFA metric is calculated by considering a prediction $y_i$ correct if it is either equal to the ground truth label $\hat{y}_i$ or a class at a one-hop distance from $\hat{y}_i$. This metric allows the model to make wrong label predictions as long as they are ordinally close to the ground truth.

\section{Results and Discussion}\label{section:results}

We evaluate SeSaME by comparing the ASR performance when fine-tuned on the predictions of the graph-semantic estimation model versus random sampling. This is done by (1) analyzing the validation metrics for GNN training on the ordinal regression task and (2) estimating the saliency of the retrieved data points by conducting three comparison fine-tuning experiments.

To this end, we define the evaluation datasets:

\begin{enumerate}
    \item \textbf{random dataset}: $10$k data points randomly sampled from the test set,
    \item \textbf{difficult dataset}: $900$ data points from the test set labelled as very difficult to learn,
    \item \textbf{semantic similarity dataset}: $2$k data points from the test set labelled as difficult, which also exhibit a high semantic correspondence with the GNN sample. More precisely, we first sample difficult points from the holdout set as identified by the GNN model. Then, we extract difficult points from the test set, and calculate the average cosine similarity between each test point against the GNN sample. Finally, we pick the first $2$k test data points with the highest similarity with the GNN sample, enforcing textual correspondence between the two datasets.
\end{enumerate}

For fine-tuning the ASR model, we sample $k$ difficult nodes as predicted by SeSaME. The sampling process involves picking the nodes with inferred labels starting from the last bucket as defined in $X_c$. If the current bucket does not cover the whole sample size $k$, we move one bucket to the left until the sample is complete. For the last bucket we access, i.e., a class with size $|x_i| > (k - \sum_{i+1}^n |x_j|)$, we randomly pick the remaining $(k - \sum_{i+1}^n |x_j|)$ utterances from $x_i$.

When evaluating against the random and semantic similarity datasets, the fine-tuning process is done on $20$ epochs. When evaluating against the difficult dataset, the fine-tuning takes place for only $10$ epochs because the models do not converge for longer training.

\textbf{GNN training } The performance of the GNN models on the ordinal regression task is summarized in Table \ref{Tab:gnn_results}. We report train and validation accuracy, and one-frame agreement (OFA). The calculated gain is between GIN and the random predictions, indicating how many times GIN is better at mapping textual information to speech recognition performance.

The results show a performance increase from the random accuracy of $14.3\%$ to a SAGE accuracy of $27.6\%$. The evaluation MSE drops from the random value of $0.40$ by $31\% - 38\%$ to an MSE of $0.25$ for SAGE, and an MSE of $0.22$ with GAT. In short, we observe that both local and non-local GNN models can successfully map a transformation between textual utterances to predicted ASR performance.

\begin{table}[!htb]
	\centering
	\begin{tabular}{lccccccccccc}
	\toprule
	\toprule
	& Train & Val & Train & Val & MSE \\
	& Acc. & Acc. & OFA & OFA & \\
	\midrule
	 GCN & 27.3\% & 27.3\% & 72.2\% & 72.3\% & 0.26 \\
	 GIN & 27.7\% & 27.7\% & 73.2\% & 73.4\% & 0.25 \\
	 SAGE & 27.4\% & \textbf{27.6}\% & 72.3\% & \textbf{72.9}\% & \textbf{0.25} \\
	 GAT & 27.7\% & \textbf{27.5}\% & 72.3\% & \textbf{73.1}\% & \textbf{0.22} \\
	 Random & - & \textbf{14.3}\% & - & \textbf{39.7}\% & \textbf{0.40} \\
    \midrule 
    \bottomrule
	\end{tabular}
	    \caption{GNN performance on bucketed WER Ordinal Regression. \smallskip}
	\label{Tab:gnn_results}
    \end{table}

\textbf{ASR fine-tuning} The random test results indicate that both the test loss and WER do not differ between sampling approaches. However, the WER drops from $18.2\%$ without fine-tuning to approximately $17\%$ when using any GNN. This indicates that when evaluating against a random test dataset, any sampling approach has the same positive effect relative to no further model fine-tuning. To compare with existing metrics, the SOTA performance on a speech recognition task using the CommonVoice English test set is reported to a WER of $14.81\%$ using a language modelling head, and WER of $19.06\%$ without the LM head. We achieved a WER of $16\%$ on the test set when fine-tuning the XLS-R-300M model for $20$ epochs with the CTC task. The similar test loss and WER results across GNN and random sampling indicate that \emph{SeSaME does not improve ASR fine-tuning when evaluated on random test data}.

\begin{table}[!htb]
	\centering
	\begin{tabular}{lccccccccccc}
	\toprule
	\toprule
	 & Train Loss & Test Loss & WER \\
	\midrule
	 GCN & 0.69 & 0.63 & 17\%\\
	 GIN & 0.72 & 0.63 & 16.9\%\\
	 SAGE & 0.66 & \textbf{0.63} & \textbf{16.9\%}\\
	 GAT & 0.51 & \textbf{0.64}  & \textbf{16.9}\%\\
	 Random x 5 & 0.66 $\pm$ 0.45-3 & \textbf{0.63}  & \textbf{17}\%\\
	 W/o fine-tuning & - & - & \textbf{18.2\%}\\
    \midrule
    \textbf{Gain} & - & - & 7.14\% \\
    \bottomrule
	\end{tabular}
			\caption{ASR fine-tuning with a sampled subset of $4k$ out of $5k$ data points; Evaluated on $10k$ random data points drawn from the test data. Fine-tuned for 20 epochs. \smallskip}
	\label{Tab:eval1_results}
    \end{table}

Table \ref{Tab:eval3_results} shows the results when evaluating against the difficult dataset. Unlike the previous experiment, this evaluation highlights a considerable difference in training loss values between GNN and the random baseline, e.g., $1.56$ for GIN, and $0.69$ for the random sample, indicating that the GNN models are indeed able to predict and sample data points that are difficult to train on. The GNN test loss values are considerably lower than random, indicating that sampling difficult points can improve the test loss, but the evaluation WER metric for local aggregation is lower than the random sampling baseline. However, GAT shows significantly better performance, with a $17\%$ decrease in test loss compared to random sampling. GAT also achieves an impressive WER decrease of $37.9\%$ compared to the random $40.9\%$, meaning that \emph{using the attention mechanism, we can effectively improve fine-tuning of an ASR system without passing the data through the model, and without any audio processing}. These results indicate that \emph{we can apply SeSaME combined with non-local aggregation to efficiently sample and fine-tune an ASR model on difficult points.}


\begin{table}[!htb]
	\centering
	\begin{tabular}{lccccccccccc}
	\toprule
	\toprule
	 & Train Loss & Test Loss & WER \\
	\midrule
	 GCN & 1.59 & 1.48 & 41.8\% \\
	 GIN & \textbf{1.56} & \textbf{1.48} & \textbf{41.6\%} \\
	 SAGE & \textbf{1.59} & \textbf{1.49} & \textbf{41.7}\%\\
  GAT & \textbf{1.39} & \textbf{1.32} & \textbf{37.9\%} \\
	 Random x 5 & \textbf{0.69} & \textbf{1.59} & \textbf{40.9\%}\\
    \midrule
    \bottomrule
	\end{tabular}
			\caption{ASR fine-tuning with a sampled subset of $900$ out of $13K$ data points; Evaluated on $900$ very difficult points drawn from the test data. Fine-tuned for 10 epochs. \smallskip}
	    \label{Tab:eval3_results}
    \end{table}

Results from Table \ref{Tab:gnn_results} clearly indicate that semantically-driven graph structures can predict the difficulty of incoming data for ASR training. However, tables \ref{Tab:eval1_results} and \ref{Tab:eval3_results} show that leveraging this information for ASR fine-tuning is non-trivial, as data points predicted as difficult do not necessarily exhibit semantic correlation to the difficult points in the test set. The reasons behind an incoming utterance being predicted as difficult can be manyfold, e.g., background noise, incomprehensible audio, or simply highly different content. To better understand the role of semantic correlation between incoming utterances and the test dataset, we conduct one final evaluation on a subset of the test data with close textual cosine similarity to the GNN sample (see table \ref{Tab:evalf_results}).

\begin{table}[!htb]
	\centering
	\begin{tabular}{lccccccccccc}
	\toprule
	\toprule
	 & Train Loss & Test Loss & WER \\
	\midrule
	 SAGE & 0.65 & 0.62 & \textbf{16.4\%}\\
	 GAT & \textbf{0.54} & 0.62 & \textbf{16.4\%}\\
	 Random & \textbf{0.67} & 0.62 & \textbf{16.7\%}\\
    \midrule
    \bottomrule
	\end{tabular}
			\caption{ASR fine-tuning with a sampled subset of $2k$ out of $5k$ data points; Evaluated on $2k$ test points with high textual correspondence with the GNN sample; Fine-tuned for 20 epochs. \smallskip}
	    \label{Tab:evalf_results}
    \end{table}

Our findings indicate that having semantic similarity between the sampled and test datasets does help in lowering the WER of non-local GNNs. Table \ref{Tab:evalf_results} shows that fine-tuning with the semantically similar dataset as sampled using GAT brings the training loss from the random baseline value of $0.67$ to a much better performance of $0.54$. Interestingly, when evaluating against difficult points (Table \ref{Tab:eval3_results}), GAT shows an impressive $7\%$ drop in WER; however, it is gained back in Table \ref{Tab:evalf_results} when we add textual correspondence. While this seems counterintuitive, evaluating against the most difficult $7\%$ of the data points (Table \ref{Tab:eval3_results}) versus the most difficult $40\%$ data points (Table \ref{Tab:evalf_results}) is a significantly heavier task. For the first one, there is more room for improvement in the model evaluation performance, as confirmed by the evaluation WER. While for the latter, fine-tuning is trivial, therefore there is little room for substantial improvement. 

\section{Conclusion}\label{section:conclusion}
We presented \textbf{SeSaME} (\textbf{SE}mantics for data \textbf{SA}liency in \textbf{M}odel performance \textbf{E}stimation), a graph-based approach for identifying salient data points based on semantic similarity. The application of this approach is studied in the use-case of low-resource ASR models, which make heavy use of speech processing and transcription calls.

To assess SeSaME, we conduct a comparative analysis of the ASR performance when fine-tuned with graph-based estimation versus random sampling. The results indicate that our method can successfully predict the speech recognition performance of the dataset based on textual information alone. Moreover, the results clearly indicate that we can use this information for efficiently fine-tuning an ASR model when combined with a robust semantic similarity relationship between the datasets and an attention mechanism. However, fine-tuning becomes a more complex problem when employing local aggregation.

\textbf{Future research } For future research, we propose three directions: (1) a more extensive study on how semantic correlation between difficult points in sampled and test data influences fine-tuning performance, (2) an approach using the full version of RHO-LOSS as described in Equation \ref{eq:rho}, and (3) including a feedback loop from the fine-tuning predictions to the graph structure for redefining the WER labels.

\small
\bibliographystyle{plainnat}
\bibliography{main}

\newpage
\appendix
\section{Notations}\label{section:appendix}

\begin{table}[!htb]
    \centering
    \caption{A list of notations}
	\begin{tabular}{ll}
	\midrule
      $\mathcal{D}_{train}$ & train dataset \\
      $\mathcal{D}_{holdout}$ & holdout dataset, i.e. new data \\
      $x$ & input audio waveform to the ASR model \\
      $y$ & ground truth for the ASR model \\
      $X_c$ & WER buckets as classes \\
      $\mathcal{L}[y|x; D]$ & utility function mapping WER for input $x$ \\ 
	 $\mathcal{V}$ & set of nodes as utterances in the train dataset\\
	 $\mathcal{E}$ & set of edges as connections in the train dataset\\
      $\mathcal{G}$ & graph structure built on the train dataset\\
	 $\mathcal{V'}$ & subset of nodes from a larger graph\\
	 $\mathcal{E'}$ & subset of edges from a larger graph\\
      $\mathcal{G'}$ & subgraph structure of the train dataset graph structure\\
	 $\bar{\mathcal{V}}$ & new set of nodes including incoming data\\
	 $\bar{\mathcal{E}}$ & new set of edges including incoming dependencies\\
      $\bar{\mathcal{G}}$ & new graph including incoming nodes\\
      $\alpha$ & ASR model \\
      $\tau$ & TTS model \\
      $\gamma$ & GNN model \\
      $\mathsf{BERT}$ & model for generating contextualized word embeddings \\
      XLS$-$R$-$300M & open-source ASR model \\
      
    \midrule
	\end{tabular}
	    \label{Tab:table_of_notations}
    \end{table}


\end{document}